\documentclass[sigconf]{acmart}

\usepackage{booktabs} 
\usepackage{wrapfig}

\usepackage[ruled]{algorithm2e} 

\SetAlFnt{\small}
\SetAlCapFnt{\small}
\SetAlCapNameFnt{\small}
\SetAlCapHSkip{0pt}

\copyrightyear{2024} 
\acmYear{2024} 
\setcopyright{rightsretained} 
\acmConference[SIGGRAPH Conference Papers '24]{Special Interest Group on Computer Graphics and Interactive Techniques Conference Conference Papers
'24}{July 27-August 1, 2024}{Denver, CO, USA}
\acmBooktitle{Special Interest Group on Computer Graphics and Interactive
Techniques Conference Conference Papers '24 (SIGGRAPH Conference Papers
'24), July 27-August 1, 2024, Denver, CO, USA}
\acmDOI{10.1145/3641519.3657491}
\acmISBN{979-8-4007-0525-0/24/07}

\begin{document}
\title[Tele-Aloha]{Tele-Aloha: A Low-budget and High-authenticity Telepresence System Using Sparse RGB Cameras}

\author{Hanzhang Tu}
\orcid{0009-0003-7555-9546}
\affiliation{%
    \institution{Tsinghua University}
    \city{Beijing}
    \country{China}}
\email{thz22@mails.tsinghua.edu.cn}

\author{Ruizhi Shao}
\orcid{0000-0003-2188-1348}
\affiliation{%
    \institution{Tsinghua University}
    \city{Beijing}
    \country{China}}
\email{shaorz20@mails.tsinghua.edu.cn}
\author{Xue Dong}
\affiliation{%
    \institution{BOE Technology Group}
    \city{Beijing}
    \country{China}}
\email{dongxue@boe.com.cn}
\author{Shunyuan Zheng}
\orcid{0000-0001-5056-614X}
\affiliation{%
    \institution{Harbin Institute of Technology}
    \city{Weihai}
    \country{China}}
\email{sawyer0503@hit.edu.cn}

\author{Hao Zhang}
\affiliation{%
    \institution{BOE Technology Group}
    \city{Beijing}
    \country{China}}
\email{zhanghao_ot@boe.com.cn}

\author{Lili Chen}
\affiliation{%
    \institution{BOE Technology Group}
    \city{Beijing}
    \country{China}}
\email{chenlili@boe.com.cn}

\author{Meili Wang}
\affiliation{%
    \institution{BOE Technology Group}
    \city{Beijing}
    \country{China}}
\email{wangml@boe.com.cn}

\author{Wenyu Li}
\affiliation{%
    \institution{BOE Technology Group}
    \city{Beijing}
    \country{China}}
\email{liwenyu_ot@boe.com.cn}

\author{Siyan Ma}
\affiliation{%
    \institution{BOE Technology Group}
    \city{Beijing}
    \country{China}}
\email{masiyan@boe.com.cn}

\author{Shengping Zhang}
\orcid{0000-0001-5200-3420}
\affiliation{%
    \institution{Harbin Institute of Technology}
    \city{Weihai}
    \country{China}}
\email{s.zhang@hit.edu.cn}

\author{Boyao Zhou}
\authornote{Corresponding authors: Yebin Liu and Boyao Zhou}
\orcid{0009-0004-4583-2676}
\affiliation{%
    \institution{Tsinghua University}
    \city{Beijing}
    \country{China}}
\email{bzhou22@mail.tsinghua.edu.cn}

\author{Yebin Liu}
\authornotemark[1]
\orcid{0000-0003-3215-0225}
\affiliation{%
    \institution{Tsinghua University}
    \city{Beijing}
    \country{China}}
\email{liuyebin@mail.tsinghua.edu.cn}

\renewcommand\shortauthors{Tu, H. et al}

\begin{abstract}
    In this paper, we present a low-budget and high-authenticity bidirectional telepresence system, Tele-Aloha, targeting peer-to-peer communication scenarios. Compared to previous systems, Tele-Aloha utilizes only four sparse RGB cameras, one consumer-grade GPU, and one autostereoscopic screen to achieve high-resolution (2048x2048), real-time (30 fps), low-latency (less than 150ms) and robust distant communication. As the core of Tele-Aloha, we propose an efficient novel view synthesis algorithm for upper-body. Firstly, we design a cascaded disparity estimator for obtaining a robust geometry cue. Additionally a neural rasterizer via Gaussian Splatting is introduced to project latent features onto target view and to decode them into a reduced resolution. Further, given the high-quality captured data, we leverage weighted blending mechanism to refine the decoded image into the final resolution of \textit{2K}. Exploiting world-leading autostereoscopic display and low-latency iris tracking, users are able to experience a strong three-dimensional sense even without any wearable head-mounted display device. Altogether, our telepresence system demonstrates the sense of co-presence in real-life experiments, inspiring the next generation of communication. 
\end{abstract}

%
%


%
%


\maketitle

\section{Introduction}

\begin{figure}
    \centering
    \includegraphics[width=0.98\linewidth]{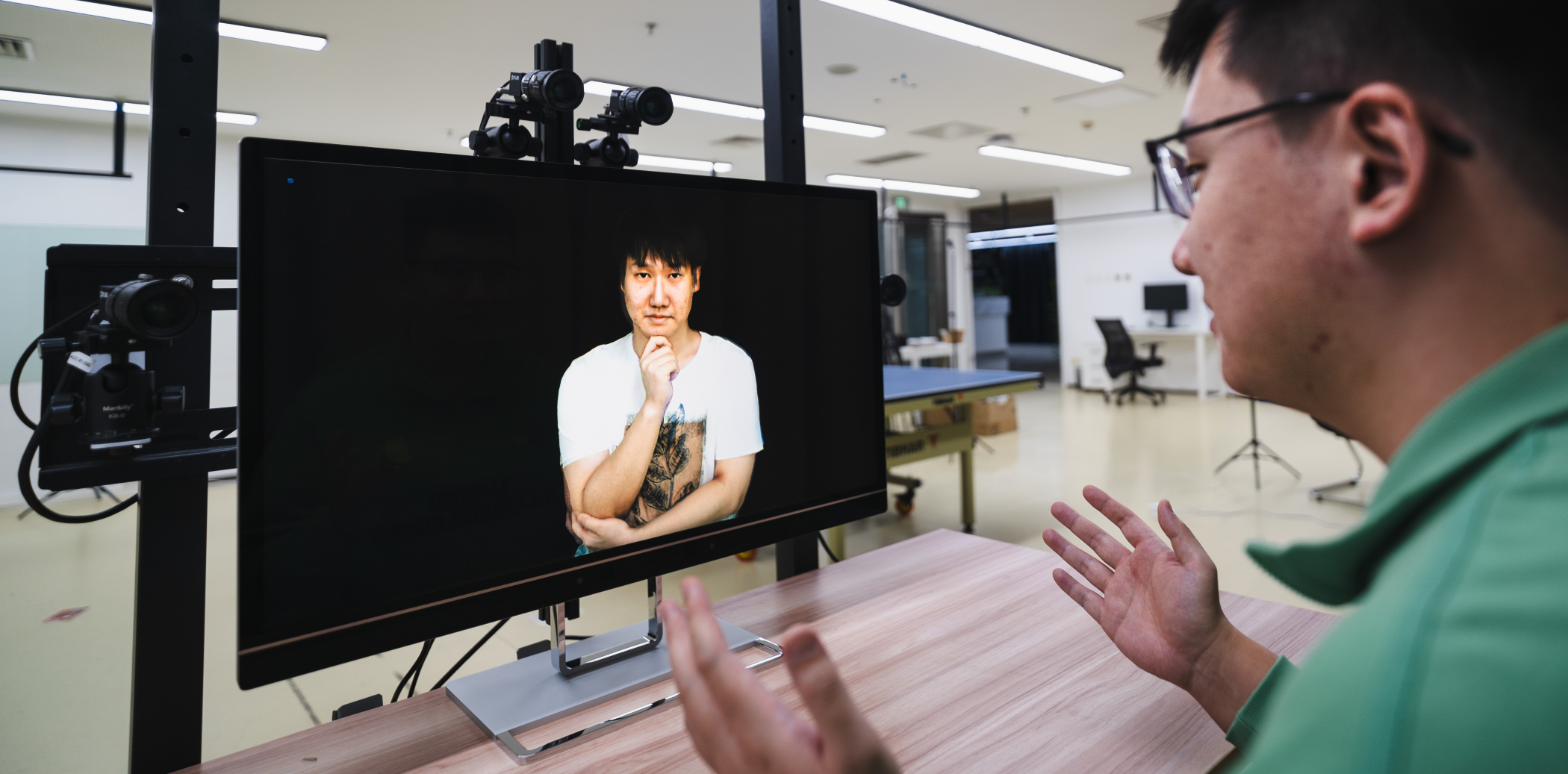}
    \caption{We design a low-budget telepresence system using only 4 consumer RGB cameras. The overall system is at an affordable price of around \$15,000. }
    \label{fig:teaser}
\end{figure}

Since ancient times, communication technology has always been one of the most important driving forces to promote innovation.
Recent years have witnessed the advancement of quality and availability of synchronous telecommunication with the help of the extraordinary development of the Internet and consumer electronics.
High-quality video conferencing systems like Zoom, FaceTime, and Teams have found their place in various scenarios, even holding dominant positions in some.
More recently, immersive telepresence systems~\cite{clemm2020toward} envisaged for 6G~\cite{strinati20196g} have attracted increasing interest due to their potential to achieve co-presence, which means individuals who are physically thousands of miles apart have the feeling of occupying a shared space.

For interpersonal communication, upper-body motion, including but not limited to facial expressions, eye gaze, hand gestures, and arm movement, dominates the body language \cite{alleva2014body, de2015perception}.
In other words, the most critical visual cues for daily communication range mainly in upper human body regions, which makes the telepresence system that specifically focuses on the upper body a more reasonable choice for research.
The latest examples such as Starline~\cite{lawrence2021project} and VirtualCube~\cite{zhang2022virtualcube} have showcased unprecedented levels of immersive experience.
However, these commercial systems necessity intricate hardware devices and customized physical setups (\textit{i.e.} booths or rooms), preventing them from spreading to average consumers.
Moreover, numerous input sensors that appear in previous systems, especially depth sensors, also increase the burden of computation and transmission, resulting in an urgent requirement for multiple high-end graphics cards.

In this paper, we present \textbf{Tele-Aloha}, \textbf{A} \textbf{lo}w-budget and \textbf{h}igh-\textbf{a}uthenticity telepresence prototype system equipped with only four RGB cameras, one consumer-grade GPU, and one autostereoscopic screen targeting upper-body peer-to-peer communication scenarios.
With this simplest hardware configuration, we are committed to enabling an immersive experience via a carefully designed system construction.
The four calibrated 4K RGB cameras, as all input sensors, capture the detailed appearance of users, which is extremely sparse compared to most of the existing systems.
In addition to the sparsity of capture devices, no depth sensor (neither TOF nor active structured light camera) appears in our system for the reason that i) they are susceptible to the environment including illumination conditions and reflection in the scene; ii) depth observations suffer from incompleteness and noise in low reflectivity areas and scenes with complex material characteristics (discussed in Sec.~\ref{subsec:cde}), resulting in a degradation of the robustness and versatility of the system.
Also, high-quality depth sensors are still costly compared to RGB cameras.
Notably, all mentioned components are readily available commercial products at affordable prices, in total around \$15,000, releasing the potential of becoming a consumer-grade product and being mass-produced.

As the core of the proposed system, we introduce a novel view synthesis algorithm, given two selected perspective views by eye tracking, to generate photo-realistic and high-fidelity rendering images under a highly sparse RGB-only setup.
Recently, Gaussian Splatting~\cite{kerbl20233d,luiten2023dynamic,wu20234d,zheng2023gps} has made a significant breakthrough in the area of novel view synthesis for its high-resolution rendering and real-time inference.
However, it typically requires per-scene parameter optimization for several minutes, which makes it unsuitable for real-world application.
In contrast, we aim to build an instant novel view synthesis system that could generalize to any unseen person through a massive human data learning process.
In our system, we focus on upper body communication, imposing a ``zoom-in'' effect of artifacts and jitters, which increases the difficulty of novel view synthesis.
Moreover, under a sparse camera setting, the wide baseline of the reference cameras also challenges the effectiveness of traditional stereo matching methods~\cite{teed2020raft,lipson2021raft}.

To this end, we carefully design a cascaded disparity estimator to obtain the geometry of the target user, which handles the problem of disparity estimation for stereo cameras with a wide baseline.
Given a more robust geometry cue, pixel-aligned features extracted from input images can be projected to the selected novel views with our generalizable 3D Gaussian Splatting rasterizer~\cite{kerbl20233d}.
We propose to splat latent features, instead of projecting spherical harmonics within 3DGS, onto the target view in a reduced resolution so that the rendering result can be further completed with a decoder network.
However, the decoded image is in a lower resolution for the reason of fast and completed rendering.
Thus, we further take advantage of the high-resolution RGB input with a weighted blending mechanism according to surface visibility in the source view, providing strong cues to a refinement module for a high-quality rendering.
The proposed algorithm balances well the trade-off between rendering quality and efficiency.
To summarize, our contributions include:

(i) A lightweight, consumer-affordable 3D telepresence system, using only four RGB cameras and no depth sensors to achieve high-resolution ($2048\times 2048$), real-time (30 fps), low-latency (less than 150 ms), and distant communication.

(ii) A cascaded stereo matching strategy to improve the robustness of depth estimation under wide baseline camera setting.

(iii) Rasterizing latent code from the source view to the target view via a generalizable Gaussian Splatting, together with a decoder network, to guarantee the rendering completeness.

(iv) Taking advantage of original high-resolution input by blending it into the novel view, along with the decoded features rasterized by 3DGS, for refinement to realize fast high-quality rendering.

\begin{table*}[htpb!]
    \caption{\textbf{3D Telepresence Systems.} Since Holoportation, MetaStream, and Live4d capture and transmit volumetric videos rather than 2D free-view videos, the resolution value cannot be quantified exactly. F.B. and U.B. stand for full-body and upper-body respectively. $^*$ LookinGood develops two systems for full-body and upper-body specifically, we only list the setting of the upper-body system.}
    \label{tab:tel-systems}
    \begin{tabular}{lcccccc}
        \toprule
        Systems                                    & Input setting   & Efficiency & Resolution         & GPU equipment                                 & Complexity    \\
        \midrule
        Holoportation~\cite{orts2016holoportation} & 8 $\times$ RGBD & 34 fps     & -                  & 10 $\times$ Titan X                           & Medium (F.B.) \\
        LookinGood$^*$~\cite{martin2018lookingood} & 1 $\times$ RGBD & 34 fps     & 512 $\times$ 1024  & 1 $\times$ Titan V                            & High (U.B.)   \\
        Starline~\cite{lawrence2021project}        & 3 $\times$ RGBD & 60 fps     & 1600 $\times$ 1200 & 2 $\times$ RTX 6000 + 2 $\times$ Titan RTX    & High (U.B.)   \\
        VirtualCube~\cite{zhang2022virtualcube}    & 6 $\times$ RGBD & 30 fps     & 1280 $\times$ 960  & 2 $\times$ RTX 3090 + 1 $\times$ RTX 2080     & High (U.B.)   \\
        MetaStream~\cite{guan2023metastream}       & 4 $\times$ RGBD & 30 fps     & -                  & 4 $\times$ Jetson Nano + 1 $\times$ RTX 2080S & Medium (F.B.) \\
        AI-mediated~\cite{stengel2023ai}           & 1 $\times$ RGB  & 24 fps     & 512 $\times$ 512   & 1 $\times$ RTX 6000 + 1 $\times$ RTX 4090     & Low (Head)    \\
        Live4d~\cite{zhou2023live4d}               & 20 $\times$ RGB & 24 fps     & -                  & 4 $\times$ RTX 3090 +  1 $\times$ RTX 4090    & Medium (F.B.) \\
        Ours                                       & 4 $\times$ RGB  & 30 fps     & 2048 $\times$ 2048 & 1 $\times$ RTX 4090                           & High (U.B.)   \\
        \bottomrule
    \end{tabular}
\end{table*}

\section{Related Work}

The 3D telepresence was identified early on \cite{raskar1998office, gibbs1999teleport} and has sparked sustained interest due to its immersive user experience.
The existing 3D telepresence systems can mainly be categorized into three types: head, upper-body, and full-body systems, as listed in Tab.~\ref{tab:tel-systems}.
The system that only focuses on the head is the simplest setting due to the highly symmetrical and non-occlusion nature of heads.
\cite{stengel2023ai} lifts a facial RGB image to 3D space using a triplane based 3D GAN~\cite{trevithick2023real}.
The full-body setting~\cite{orts2016holoportation, guan2023metastream, zhou2023live4d} poses challenges of self-occlusion issues.
Nonetheless, the small effective proportion of humans within image space results in a higher tolerance for blurry or ghosting artifacts.
We target an upper-body telepresence system similar to Starline~\cite{lawrence2021project} and VirtualCube~\cite{zhang2022virtualcube}, which satisfies an ideal communication scenario covering motion and gesture.
However, complicated upper-body movements introduce severe self-occlusion in this context, making it more difficult.
%

The primitive telepresence systems~\cite{gibbs1999teleport,kuster2012towards} attempt to capture the appearance and geometry of participants, thus they typically employ depth sensors~\cite{izadi2011kinectfusion, newcombe2015dynamicfusion} to serve as a geometry proxy.
Leveraging depth sensors, \cite{novotny2019perspectivenet,nguyen2022free,neff2021donerf,stelzner2021decomposing} facilitate and speed up the rendering process.
For example, some attempts\cite{maimone2012enhanced, newcombe2015dynamicfusion, yu2018doublefusion, su2020robustfusion} directly obtain textured mesh via depth triangulation or depth fusion to demonstrate 3D surface.
With the development of learning methods, depth maps can be refined, inpainted, and denoised with neural networks and serve as geometry to accelerate the neural rendering process.
Function4d~\cite{yu2021function4d} integrates dynamic fusion and implicit surface reconstruction to perform real-time full-body human volumetric capture from four consumer RGBD sensors.
Equipped with similar sensors, VirtualCube~\cite{zhang2022virtualcube} builds real-world cubicles to address immersive 3D video conferences under three various scenarios.
HVS-Net~\cite{nguyen2022free} warps latent feature of input view image to target view by using un-projected points of depth map, the feature is then decoded and refined with CNN based network.
LookinGood~\cite{martin2018lookingood} employs volumetric fusion~\cite{curless1996volumetric} of depth inputs to generate geometry prior for novel view rendering, but accumulative geometry error is inevitable in fusion methods when increasing image range.
FWD~\cite{cao2022fwd} refines the depth value of a captured map and warps input images into the target view.
Since the captured depth map plays the role of geometry cue, the rendering results highly depend on the quality of the captured depth map, which is sensitive to the lighting environment.

Given the increasing resolution of RGB cameras, more real-world applications utilize only RGB cameras as input to generalize to different lighting environments.
Image-based rendering (IBR)~\cite{hedman2018deep, riegler2021stable} synthesizes novel views by reasoning a blending weight and a geometry proxy, which are used to warp source image cues to novel viewpoints.
As for human novel view synthesis, NeuralHumanFVV~\cite{suo2021neuralhumanfvv} proposes a neural blending mechanism to conduct image warping and texture blending based on a neural geometry reconstruction~\cite{saito2019pifu} from sparse views.
Floren~\cite{shao2022floren} realizes a real-time full-body $360^\circ$ free-view rendering system with eight RGB cameras.
Neural radiance field (NeRF)~\cite{mildenhall2020nerf,zhang2020nerf++,barron2022mip} has shown impressive results in 4D performance capture~\cite{pumarola2021d,shao2023tensor4d}.
However, such methods typically require per-scene optimization, which restricts their applications in real-time telepresence systems.
The follow-up work~\cite{yu2021pixelnerf, wang2021ibrnet, lin2022efficient} combines the advantages of IBR and NeRF by replacing the input of implicit function from scene-specific position encoding to image features aggregated from source views.
Despite the progress in generalization, these methods rely on dense sampling points and still have difficulty achieving photorealistic results.
Recently, 3DGS~\cite{kerbl20233d, luiten2023dynamic} has introduced a new promising point-based representation.
It demonstrates a more reasonable mechanism for back-propagating the gradients alongside real-time rendering efficiency.
However, the original 3DGS requires scene-specific training in minutes.
Concurrent work~\cite{zheng2023gps, charatan2023pixelsplat, szymanowicz24splatter} attempt to address this fragmentation by formulating 3DGS on 2D image planes, termed Gaussian maps.
The Gaussian maps trained on large-scale images determine the parameters of 3D Gaussians in a feed-forward manner rather than an iterative optimization way, which makes the representation generalizable to the domain of training data.
GPS-Gaussian~\cite{zheng2023gps} is the most related method to ours.
Although they have showcased remarkable results in full-body cases, they have no specific design for the complicated self-occlusion. We summarize the major distinctions as follows.
First, compared with GPS-Gaussian, cameras and subjects are closer in our system, which leads to much larger disparity. To address this issue, we propose cascaded disparity estimation for upper-body setup to overcome the significantly larger parallax (Fig. \ref{fig:casabla}). Second, we propose to splat latent features followed by a neural decoder instead of RGB values used in GPS-Gaussian, to mitigate the incompleteness caused by severe self-occlusion (Fig. \ref{fig:comp}). Third, we propose a lightweight refinement module that improves rendering quality without introducing much computational burden (Fig. \ref{fig:abla-blending}).

\section{System Overview}

\begin{figure*}
    \includegraphics[width=\linewidth]{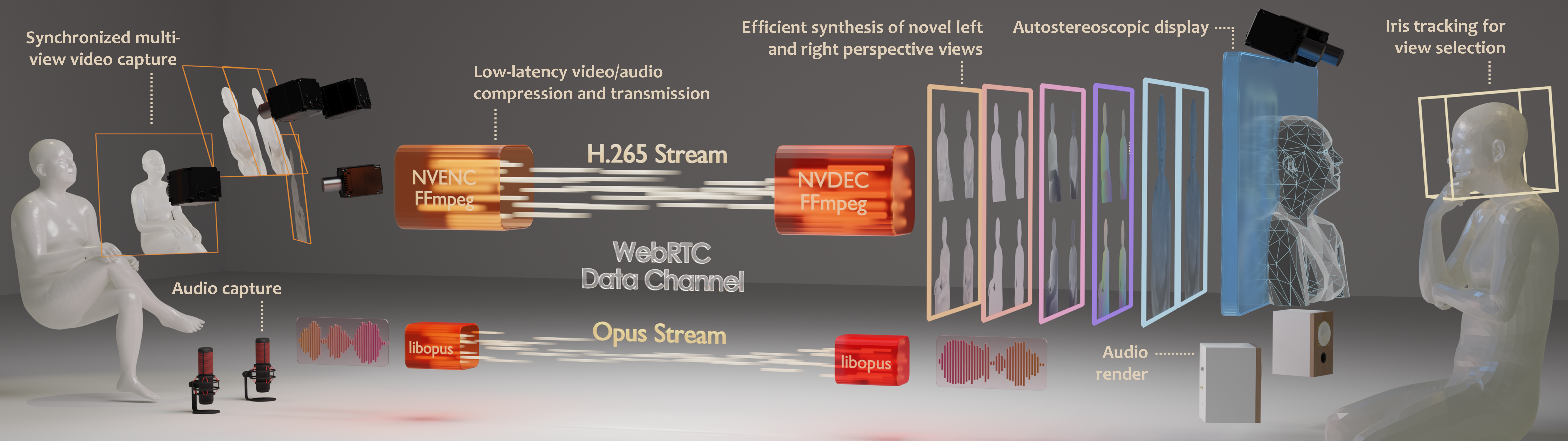}
    \caption{Schematic diagram of the overall system.}
    \Description[sys model]{System model}
    \label{system-model}
\end{figure*}

The schematic diagram of our system is shown in Fig. \ref{system-model}.
We construct the hardware and software architecture with concentrated attention to optimized user experience.
In general, our system consists of video capture, transmission, stream decoding, novel view synthesis, and display components.

\subsection{Overall system setup}

In our design, we are committed to providing users with a seamless and unencumbered communication experience.
The users are accommodated in front of the display with a typical 1.25m eye-to-display distance. The virtual remote user is rendered with a $180^{\circ}$ rotation about the eye-to-display midpoint as described in \cite{lawrence2021project}.
To include important non-verbal cues like hand gestures on a relatively smaller screen size (27-inch in our implementation), we apply a scaling factor of 0.5 for rendering the upper body. To avoid vergence-accommodation conflict and maximize the user's comfort, we choose to put the virtual user near the display plane rather than behind the screen. The setup may break the eye contact to some extent, so we use the center point of the two eyes as the scaling center to minimize the defocus of eye contact.

During conversation, the optimal range that the user can move left and right is about 0.8m, which means the view angle is about 40°. The range of forward and backward movement is about 0.3m. The limitation is mainly due to the characteristics of the display hardware and the capture volume of cameras.

\subsection{Multiview capture system}

Our system captures 4 video streams using 4 RGB cameras.
Since we get rid of costly and illumination-sensitive depth sensors, we propose to apply stereo matching~\cite{lipson2021raft} of dual-camera setting as a fast and accurate geometry proxy.
As shown in Fig~\ref{system-model}, we place four hardware-synchronized RGB cameras on the left, right, and above the displayer to provide complete coverage of the user.
Each camera(BFS-U3-123S6C-C) has a high resolution of $4096\times 3000$ and operates at 30 Hz.
Two upper cameras make up a dual-view stereo with a narrow baseline, while the other pair has a wider one.
The former provides accurate geometric information to the latter as an initialization for robust depth estimation.
Notice that although we choose cameras with global shutter in our implementation, our system is not specifically designed for such cameras. Rolling shutter is also feasible when subject motion is not extremely fast.

Precise geometric knowledge of the system including camera-camera and camera-display poses is required for
stereo matching and novel view synthesis. We calibrate cameras adopting the
method proposed by \cite{cam-calibration-zhang}, which minimized reprojection error
over images of a planar checkerboard-like target.
Also, bundle adjustment is utilized for overall refinement, providing the final
estimation of camera intrinsics and extrinsics.
With respect to camera-display calibration, method described in \cite{hesch2010mirror} is adopted to solve relative transform between the display and cameras.
In addition to camera geometries, we also color-calibrate four RGB
cameras with a standard ColorChecker. We counteract the effects of ambient lighting and
color characteristics by adjusting gamma and a $3 \times 3$ color correction matrix (CCM).
The color consistency between cameras ensures the robustness and fidelity of subsequent algorithms.

\subsection{Data compression and transmission}

Captured data will be encoded into H.265 streams and transmitted over the Internet using low-latency WebRTC technology~\cite{johnston2012webrtc}.
Hardware-based encoder and decoder \cite{nvcodec} are deployed for data stream encoding and decoding.
With end-to-end encoding/decoding offloaded to NVENC/NCDEC, the graphics/CUDA cores and the CPU cores are free for other operations.
Before encoding, all four input images are square-cropped and concatenated into a large one,
resulting in a single input image into NVENC with a resolution of $6000 \times 6000$.
As for audio, capture is performed in two-channel stereo, 16-bit and 48000 Hz.
Timestamps are also inserted into the stream for audio-video synchronization.
We measured an overall network bandwidth of 100 Mbit/s which is feasible for most enterprise or even in-home users.

\subsection{Novel view synthesis and display}

On receiving a new frame from remote client, a video matting module \cite{lin2022robust} is engaged as a pre-processing step.
Meanwhile, we track the 3D eye positions of the users, providing the parameters of the viewpoints for our algorithm to generate novel views.
For eye tracking, two of the existing camera views on the top are used as input (cam0 and cam1 in Fig. \ref{fig:algopipe}) and we use BlazeFace [Bazarevsky et al. 2019] finetuned with our own data to detect 2D iris positions, which are further triangulated into 3D positions.
Given foreground images and tracked eye positions, the novel view synthesis (elaborated in Sec. \ref{sec:nvs}) is performed.

As for the display, we prefer an autostereoscopic screen powered by eye tracking for its full-scale display field and much higher resolution with respect to the quilt format in light field holography device~\cite{lookingglass}, displaying all views from leftmost to rightmost into one single image.
In theory, all autostereoscopic screens satisfy our system, \textit{e.g.} 32-inch 3D display screen of Beijing Shiyan Technology Co., Ltd or \textit{ThinkVision-27-3D}~\cite{ThinkVision3d} with a cost lower than \$3,000.

With the highly-optimized, well-implemented pipeline, our system has the capability of handling all workloads with as little as one consumer-grade GPU (NVIDIA RTX 4090) at over 30 FPS.
This allows us to well organize the input and output of our novel view synthesis algorithm.
We achieve an end-to-end latency of less than 150 ms in our prototype system, where two terminals are in the same LAN, which gives good interactivity to the participants.

\section{Efficient novel view synthesis} \label{sec:nvs}

\begin{figure*}
    \centering
    \includegraphics[width=0.9\linewidth]{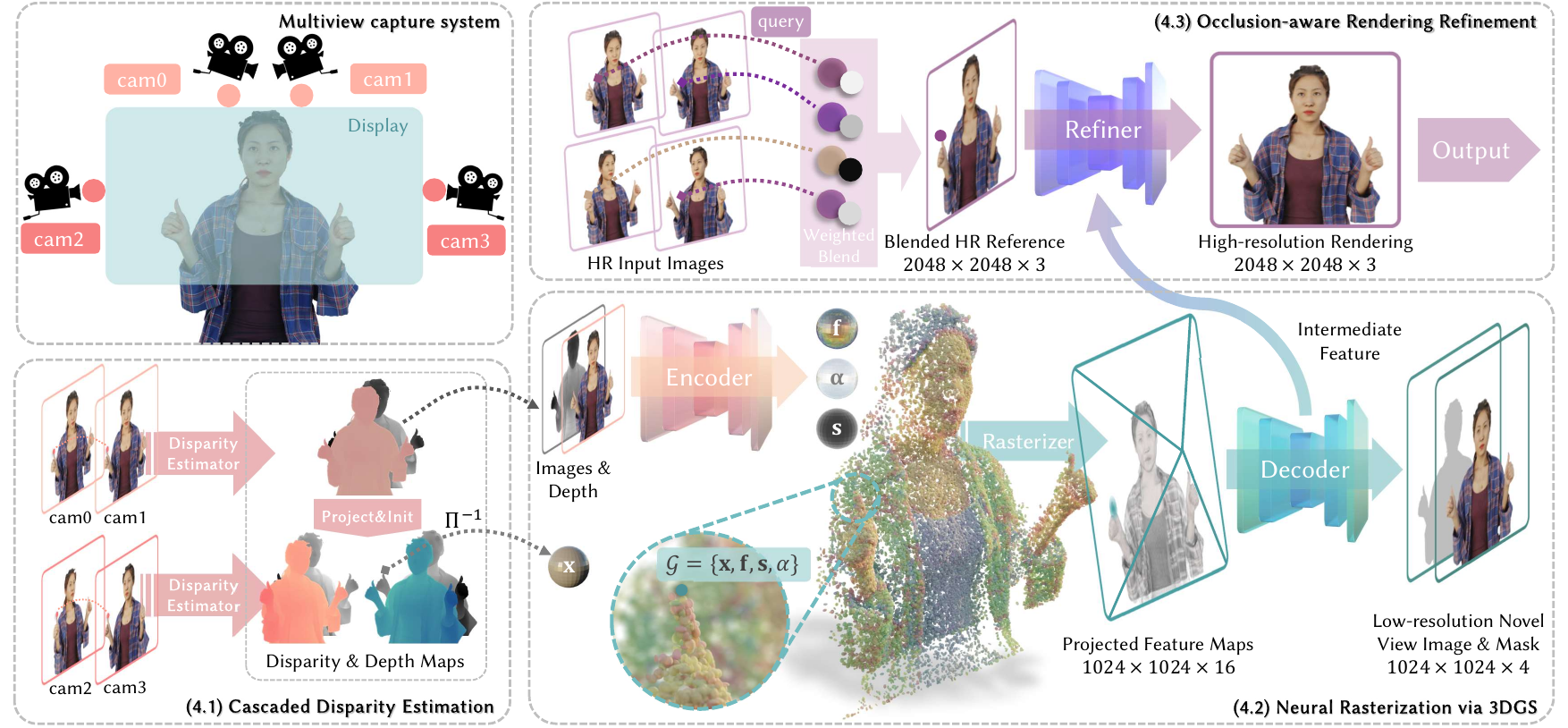}
    \caption{The pipeline of the proposed novel view synthesis algorithm.
        Given four input views that form two stereo pairs, we first estimate the disparity maps in the perspectives of these input views in a cascaded manner.
        Then, together with feature maps, scale maps, and opacity maps extracted by an image encoder, the pixels are lifted to 3D to form pixel-aligned 3DGS points.
        After that, feature vectors are rasterized to selected novel views, followed by a decoder that turns feature maps back into RGB images.
        Finally, a refiner module is applied to upsample the novel view rendering. }
    \label{fig:algopipe}
\end{figure*}

In this section, we present an efficient novel view synthesis method for the upper body with only four RGB images.
Our algorithm achieves photo-realistic novel view generation in real time.
Instead of solving the highly ill-posed problem of geometry reconstruction,
we turn to stereo constraint across input views and novel-view-centered neural rendering.
All computationally intensive neural networks focus on targets in the 2D domain and thus can be efficiently executed.
The pipeline of the proposed method is illustrated in Fig.~\ref{fig:algopipe}.
We first introduce a \textbf{Cascaded Disparity Estimation} (Sec.~\ref{subsec:cde}), in which the disparity of two closer cameras is predicted and serves as an initialization of the disparity of two farther ones.
In Sec.~\ref{subsec:fp}, we propose a \textbf{Neural Rasterizer}, which projects latent features from source views to the target view via Gaussian Splatting, and a decoder network for a completed rendering in reduced resolution.
Further, \textbf{Occlusion-aware Rendering Refinement} is presented in Sec. \ref{subsec:orr}, which fuses low-resolution rendering with weighted blending images from high-resolution input views, to synthesize high-quality images
in resolution of $2048 \times 2048$ less than 25ms with a single NVIDIA RTX 4090.

\subsection{Cascaded Disparity Estimation} \label{subsec:cde}

Obtaining a stable geometry is crucial for the robustness of novel view synthesis, especially in the case of sparse input views (4 views in our system). To address this, we design a cascaded multi-view capture system as illustrated in Fig.~\ref{fig:algopipe}. Two pairs of horizontally arranged cameras with different lengths of baseline are present in our system, denoted as upper pair (\textit{cam0, cam1}) and lower pair (\textit{cam2, cam3}), respectively. The upper pair has a smaller baseline, which is more robust for stereo matching, while the lower pair has a larger field of view. Therefore, we propose to estimate the disparity of the upper pair first and adopt it as initialization to estimate the disparity of the lower pair with more stability.

Specifically, we adopt RAFT-stereo~\cite{lipson2021raft} to estimate a disparity $\hat{d}_{u}$ for each pixel $\left(u, v\right)$ in reference view that matches its counterpart $(u+\hat{d}_{u}, v)$ in the target view, which is formulated as
\begin{equation}
    \langle \hat{\mathbf{d}}_1, \hat{\mathbf{d}}_2 \rangle = \mathcal{F}_{d}\left(\mathbf{I}_{1}, \mathbf{I}_{2}, \mathbf{d}_1^{init}, \mathbf{d}_2^{init}\right),
    \label{eq:disp_module}
\end{equation}
where $\mathbf{d}_1^{\text{init}}, \mathbf{d}_2^{\text{init}}$ are disparity initialization for the iterative update process.
We first apply the module $\mathcal{F}_{d}$ in Eq.~\ref{eq:disp_module} to upper pair with zero initialization
\begin{equation}
    \langle \hat{\mathbf{d}}_{cam0}, \hat{\mathbf{d}}_{cam1} \rangle = \mathcal{F}_{d}\left(\mathbf{I}_{cam0}, \mathbf{I}_{cam1}, \mathbf{0}, \mathbf{0}\right).
\end{equation}
The predicted disparity of \textit{cam0} is converted to a depth map and then lifted to a 3D point cloud.
These points are rasterized to the viewpoints of \textit{cam2} and \textit{cam3}. Z-buffers are extracted and converted back to disparity maps $\mathbf{d}_{cam\{2, 3\}}^{\text{init}}$.
Disparity maps of lower pair can now be estimated with more robustness and stability
\begin{equation}
    \langle \hat{\mathbf{d}}_{cam2}, \hat{\mathbf{d}}_{cam3} \rangle = \mathcal{F}_{d}\left(\mathbf{I}_{cam2}, \mathbf{I}_{cam3}, \mathbf{d}_{cam2}^{\text{init}}, \mathbf{d}_{cam3}^{\text{init}}\right) .
\end{equation}

As shown in Fig.\ref{fig:algopipe}, three disparity maps $\hat{\mathbf{d}}_{\{cam0, cam2, cam3\}}$ are converted to depth maps $\hat{\mathbf{z}}_{\{cam0, cam2, cam3\}}$, and further transformed into Gaussian Splatting point cloud in neural rasterization.

\begin{figure}
    \centering
    \includegraphics[width=0.9\linewidth]{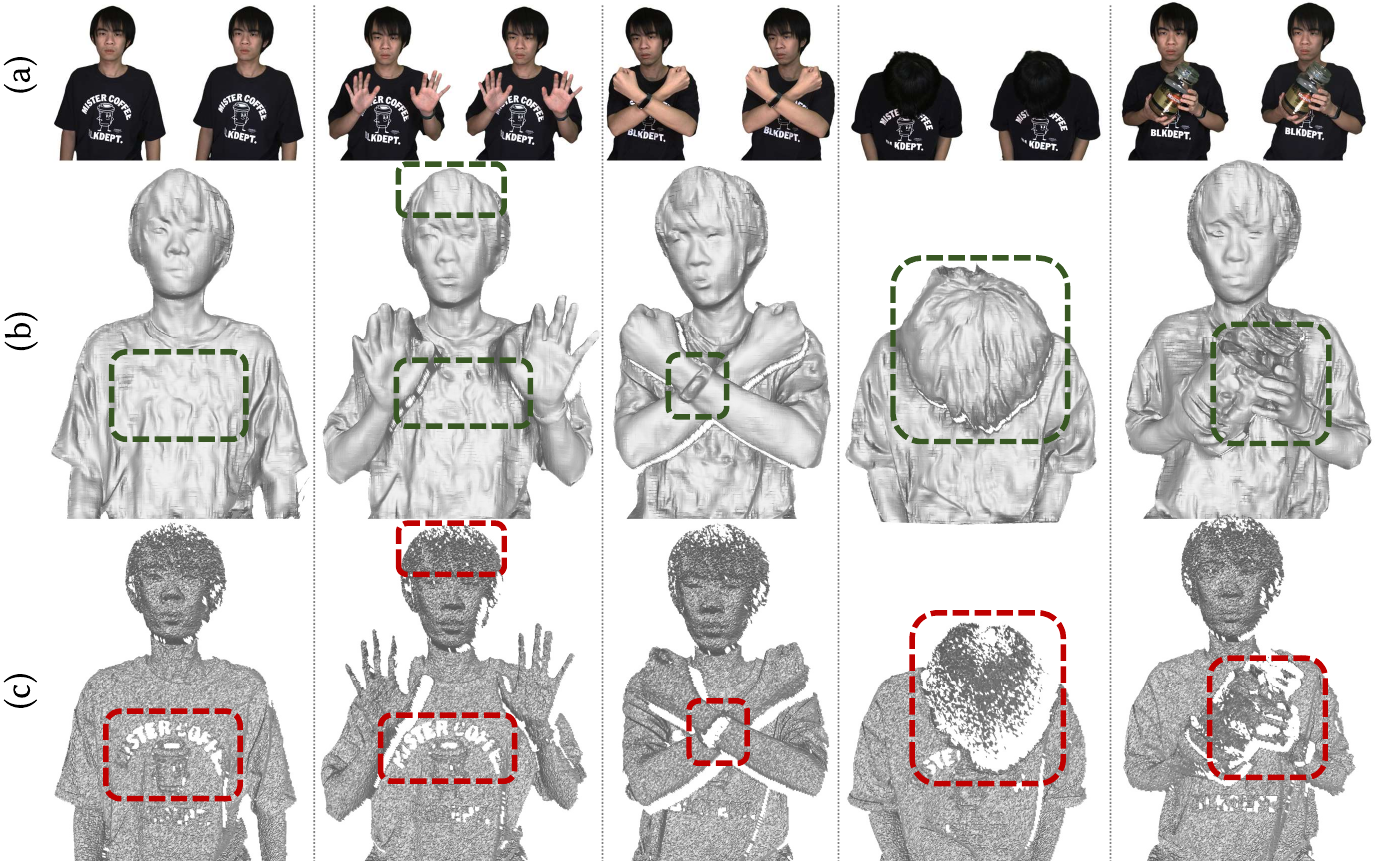}
    \caption{Comparison of depth map sources. (a) Stereo rectified pair input to the disparity estimator; (b) Depth maps predicted by the disparity estimator mentioned in Sec. \ref{subsec:cde}; (c) Depth maps captured by a TOF sensor. }
    \label{fig:depthcomp}
\end{figure}

The visualization of the depth reconstruction is demonstrated in Fig.\ref{fig:depthcomp}, which shows that using only RGB cameras (second row), our method produces competitive depth maps in comparison to that captured by a TOF sensor (third row) which costs up to 1500 USD (Lucid Helios 2~\cite{lucidhelios}).
Notably, depth observations from the TOF sensor are incomplete in areas like the logo on clothes, hair and glass jar.

\subsection{Neural Rasterization via 3DGS} \label{subsec:fp}
To synthesize novel views, we propose to transform three depth maps $\hat{\mathbf{z}}_{\{cam0, cam2, cam3\}}$ into a latent-based 3D Gaussians and train a pixel-wise image encoder $\mathcal{E}_{src}$ with a neural image decoder $\mathcal{D}_{novel}$. The image encoder predicts the Gaussian properties instantly. Then, the latent-based 3DGS is rendered as the latent map at the novel view, and the neural image decoder is utilized to recover RGB images.

\paragraph{Latent-based Gaussian Splatting.} In the latent-based Gaussian Splatting, each point $p_i$ is characterized by four properties: position $\mathbf{x}_i$, latent appearance feature $\mathbf{f}_i$, scale $\mathbf{s}_i$, opacity $\alpha_i$. The rotation of Gaussian points is set to an identity matrix. Compared to the original Gaussian Splatting, we propose to learn the latent feature instead of RGB values. Our method compresses local content information from the original image into the high-dimensional latent code. Consequently, each point can effectively represent a local region, enhancing rendering robustness against occlusion, depth estimation errors, and camera color noise.

Specifically, Given camera parameters $\Pi$, 3D positions $\mathbf{x}$ of Gaussian points can be unprojected from depth maps.
Then, the pixel-wise image encoder $\mathcal{E}_{src}$ is constructed to extract image features from source views.
After a shared backbone, three different heads are applied to the intermediate feature,
producing appearance feature $\mathcal{M}_{f}\in \mathbb{R}^{H\times W \times d}$, scale map $\mathcal{M}_s\in \mathbb{R}^{H\times W \times 3}$ and opacity map $\mathcal{M}_\alpha \in \mathbb{R}^{H\times W \times 1}$, respectively.
To emphasize the geometric awareness for more accurate Gaussian parameter regression, depth maps are also introduced to the network
\begin{equation}
    \langle\mathcal{M}_{f},\mathcal{M}_s,\mathcal{M}_\alpha\rangle=\mathcal{E}_{src}(\mathbf{I} \oplus \mathbf{z}).
\end{equation}

As illustrated in Fig.~\ref{fig:algopipe}, foreground pixels in source views (\textit{i.e.} \textit{cam0}, \textit{cam2}, and \textit{cam3}) are lifted to pixel-aligned Gaussian points $\{\mathcal{G}\}$.
These Gaussian points are then differentiably rasterized to the targeted novel views (left and right perspective of the user) given their projection matrix $\Pi_{novel}$: $\mathbf{F}_{novel} = \mathcal{R}_{\mathcal{G}} (\{\mathcal{G}\}, \Pi_{novel}), $
where $\mathcal{R}_{\mathcal{G}}$ denotes the differentiable Gaussian splatting rasterizer \cite{kerbl20233d}.
The projected feature maps $\mathbf{F}_{novel}$ carry abundant appearance details from input images.

\paragraph{Neural Image Decoder.} \label{subsec:imgdec}
Due to sparsity caused by self-occlusion, the projected feature maps are not necessarily dense. Therefore, a deep neural image decoder $\mathcal{D}_{novel}$ implemented as a 2D UNet is introduced to address the artifacts and discontinuities.
Since the encoder and the decoder are both trained on a large-scale dataset, the decoder is capable of inpainting invisible regions of the novel view from the latent map learned by the encoder.

Considering runtime efficiency and the receptive field of the network, the image decoder generates images with a reduced resolution ($1024\times 1024$, while the full output resolution is $2048\times 2048$) to enhance the completeness of the inpainted images.
Also, the refined feature maps $\mathbf{F}_{novel}^{ref}$ from the last convolutional layer are extracted for further upsampling operation:
\begin{equation}
    \langle \mathbf{F}_{novel}^{ref}, \mathbf{I}_{novel}^{lr} \rangle = \mathcal{D}_{novel} (\mathbf{F}_{novel}).
\end{equation}

We apply the perceptual loss to $\mathbf{I}_{novel}^{lr}$ for better convergence due to a shorter gradient chain compared with the final full-resolution rendering produced in Sec.~\ref{subsec:orr}.

\subsection{Occlusion-aware Rendering Refinement} \label{subsec:orr}

\begin{figure}
    \centering
    \includegraphics[width=0.95\linewidth]{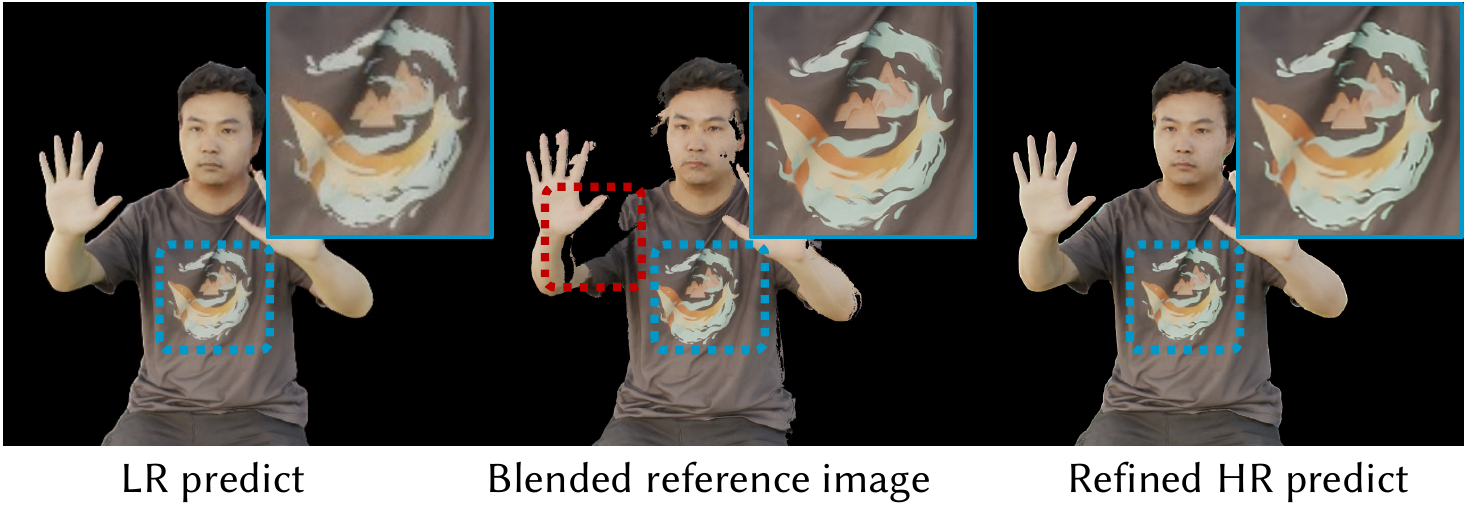}
    \caption{ Qualitative evaluation of refinement module. LR image (left) lacks details in \textcolor[RGB]{67,178,193} {blue square} while the blended image (middle) suffers from incompleteness in \textcolor{red}{red square}. The final result (right) takes the advantage of both. }
    \label{fig:abla-blending}
\end{figure}

We have now obtained a complete and hole-free novel view rendering, but there is still a gap between satisfaction in terms of resolution.
A lightweight refinement module $\mathcal{D}_{refine}$ is engaged to generate the final output in full resolution.
To fully leverage the high-resolution input of the system, $\mathcal{D}_{refine}$ takes the refined image feature $\mathbf{F}_{novel}^{ref}$ and a high-resolution image $\mathbf{I}_{blend}$ blended from source views as input, to produce high-resolution images.
As shown in Fig.~\ref{fig:abla-blending}, the blended image $\mathbf{I}_{blend}$ surpasses the inpainted image $\mathbf{I}_{novel}^{lr}$ in respect of high-frequency texture.
Meanwhile, it is not necessarily complete due to self-occlusion and inaccurate geometry prediction. Therefore, fusing the two can give full play to their complementary advantages.

\begin{figure}
    \centering
    \includegraphics[width=0.95\linewidth]{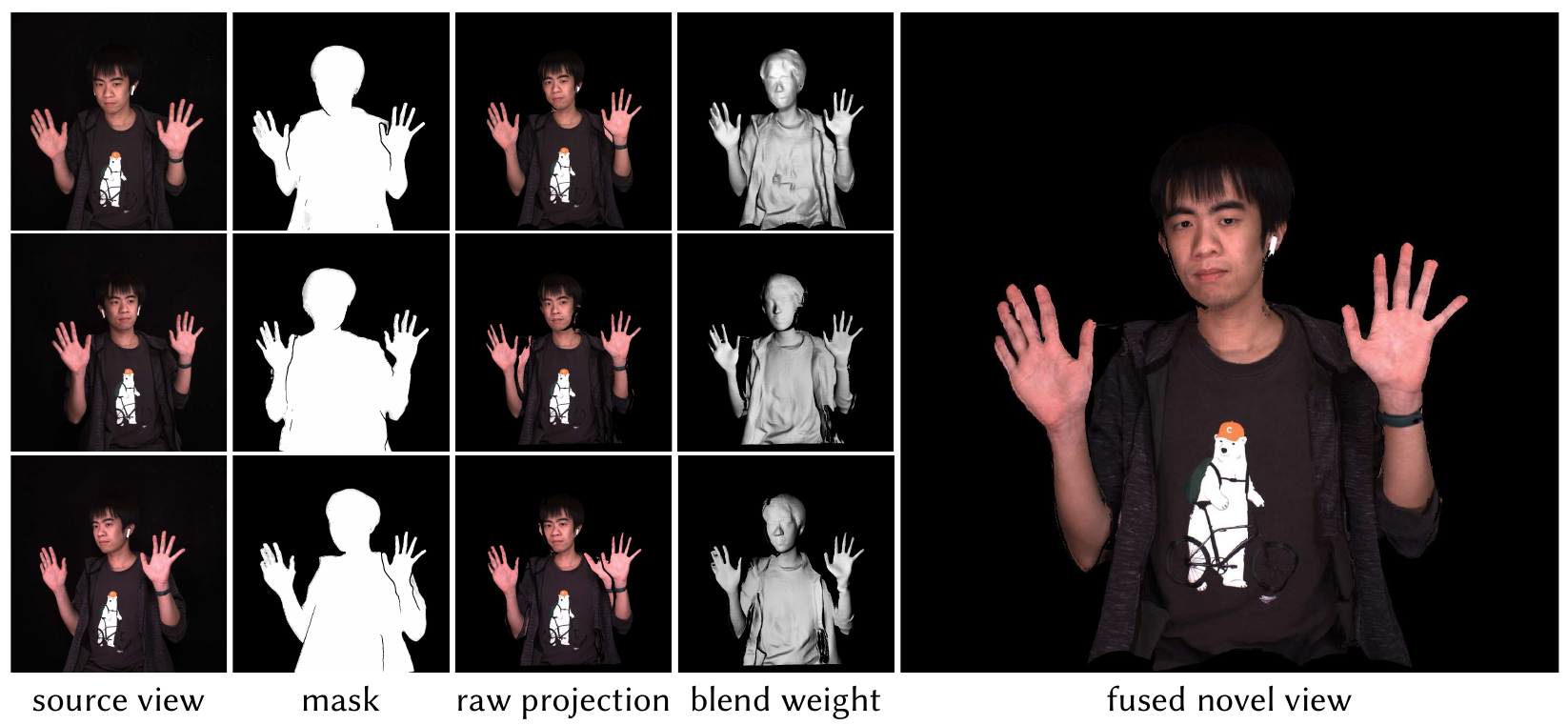}
    \caption{Source views are blended in novel view to acquire high-frequency texture with occlusion awareness. The blending weights are basically defined by surface normal, view direction, and PSDF value. }
    \label{fig:image-blending}
\end{figure}

We first introduce the definite process of input views blending, which is briefly illustrated in Fig. \ref{fig:image-blending}.
To begin, we warp all predicted depth maps $\hat{\mathbf{z}}_{i}$ in Sec.~\ref{subsec:cde} onto the novel view, resulting in a relatively dense fused depth map denoted as $\mathbf{z}_{fused}$.
After that, each pixel in $\mathbf{z}_{fused}$ is projected back to all input views:
\begin{equation}
    \mathbf{x}_i = \Pi_{i}(\Pi_{novel}^{-1}(u,v,\mathbf{z}_{fused}(u, v))).
\end{equation}
Color value and depth value are fetched from each input view using the back-projected coordinate:
\begin{equation}
    \mathbf{c}_i = \operatorname{Interp} (\mathbf{I}_i, \mathbf{x}_{i.uv}), \quad
    z_i = \operatorname{Interp} (\hat{\mathbf{z}}_{i}, \mathbf{x}_{i.uv}),
\end{equation}
where $\operatorname{Interp}(\cdot)$ is a bi-linear sampling function.
Similar to \cite{lawrence2021project}, the blending weights of color values from different input views are given by:
\begin{equation} \label{eq:blend-weight}
    w_i = \mathbf{M}_i (\mathbf{x}_{i.uv}) \cdot \operatorname{Occ} (\mathbf{x}_i) \cdot \cos \langle \mathbf{r}(u, v), \mathbf{n}_i \rangle \cdot \frac1{\|\mathbf{x}_i^{cam}\|_2},
\end{equation}
where $\mathbf{M}_i$ is a binary mask labeling valid pixels in input view $i$, which is obtained by an AND operation of the input mask, appearance consistency check mask, and edge mask,
$\operatorname{Occ}(\cdot)$ is the signed distance function (SDF) check function to avoid sampling on the occluded regions
\begin{equation}
    \operatorname{Occ} (\mathbf{x}_i) =
    \begin{cases}
        1 & \text{ if } |\mathbf{x}_{i.z}-z_i| < \delta \\
        0 & \text{ otherwise. }
    \end{cases},
\end{equation}
$\cos \langle \mathbf{r}(u, v), \mathbf{n}_i \rangle$ evaluates the angle between ray direction corresponded to pixel $(u, v)$ and the surface normal of input view,
and $\mathbf{x}_i^{cam}$ denotes the coordinate of the point in $i^th$ camera's coordinate system.
Finally, all $\mathbf{c}_i$ are weighted-summed:
\begin{equation}
    \mathbf{I}_{blend}(u, v)=\frac1{\sum_{i=0}^{\upsilon_{num}}w_{i}}\cdot\sum_{i=0}^{\upsilon_{num}}w_{i}\cdot \mathbf{c}_{i}.
\end{equation}

Finally, we fuse the two components together:
\begin{equation}
    \mathbf{I}_{novel}^{hr} = \mathcal{D}_{refine} (\mathbf{F}_{novel}^{refine} \oplus \mathcal{E}_{hr}(\mathbf{I}_{blend})),
\end{equation}
in which $\mathcal{E}_{hr}$ is an encoder composed of several edge-enhanced diverse branch blocks \cite{Wang2022CVPR}, $\oplus$ stands for concatenation. $\mathbf{I}_{novel}^{hr}$ is the overall synthesized high-fidelity novel view image.

\begin{figure*}
    \centering
    \includegraphics[width=0.95\linewidth]{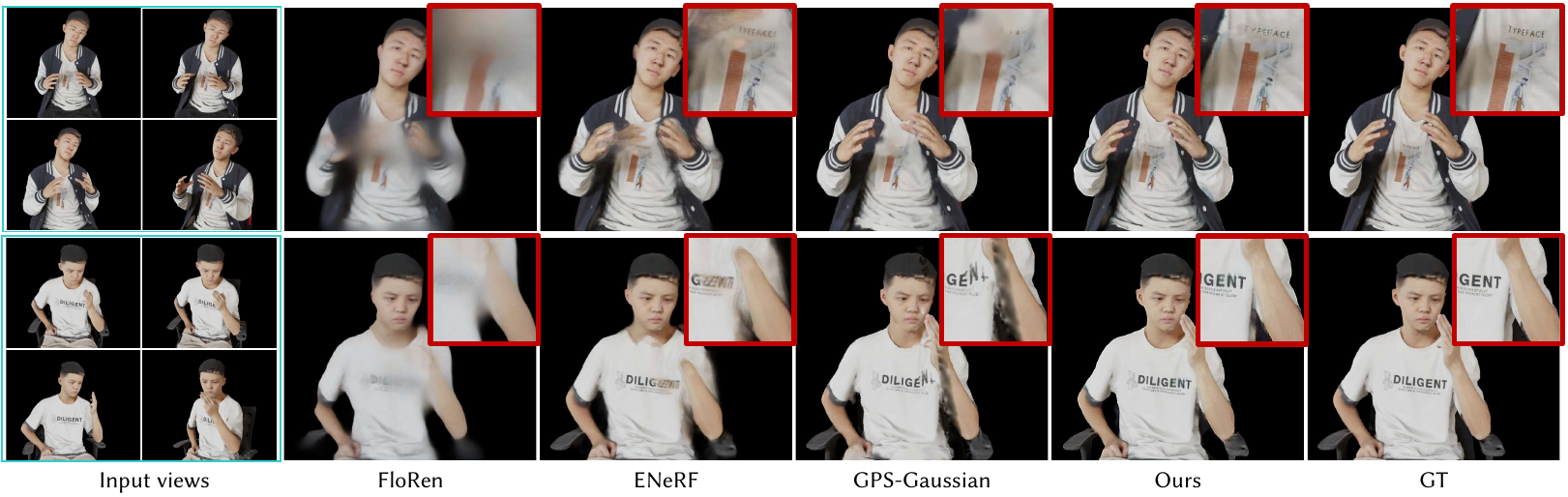}
    \caption{ Comparison on sparse-view synthetic dataset against Floren~\cite{shao2022floren}, ENeRF~\cite{lin2022efficient}, and GPS-gaussian~\cite{zheng2023gps}. }
    \label{fig:comp}
\end{figure*}

\section{Experiment and Analysis}

The proposed method is trained on THuman-Sit~\cite{zhang2023ins} and THuman2.0~\cite{yu2021function4d}. We split the THuman-Sit scans into two parts, 4000 for training and 700 for testing, according to \cite{zhang2023ins}.
We render source images parallel to our hardware setup and render novel views randomly among them.

\subsection{Analysis of Rendering Quality}

The quantitative results on the synthetic dataset are listed in Tab.~\ref{tab:comp-nvs}. Our method outperforms other efficient NVS algorithms. The qualitative comparisons are presented in Fig.~\ref{fig:comp}. The first row shows that our method inpaints high-quality results in the occlusion region, demonstrating the effectiveness of our proposed neural rasterization. As shown in the second row, our method achieves superior rendering quality thanks to the robust cascade disparity estimation.

\subsection{Analysis of Depth Reconstruction}

\begin{figure}
    \centering
    \includegraphics[width=0.95\linewidth]{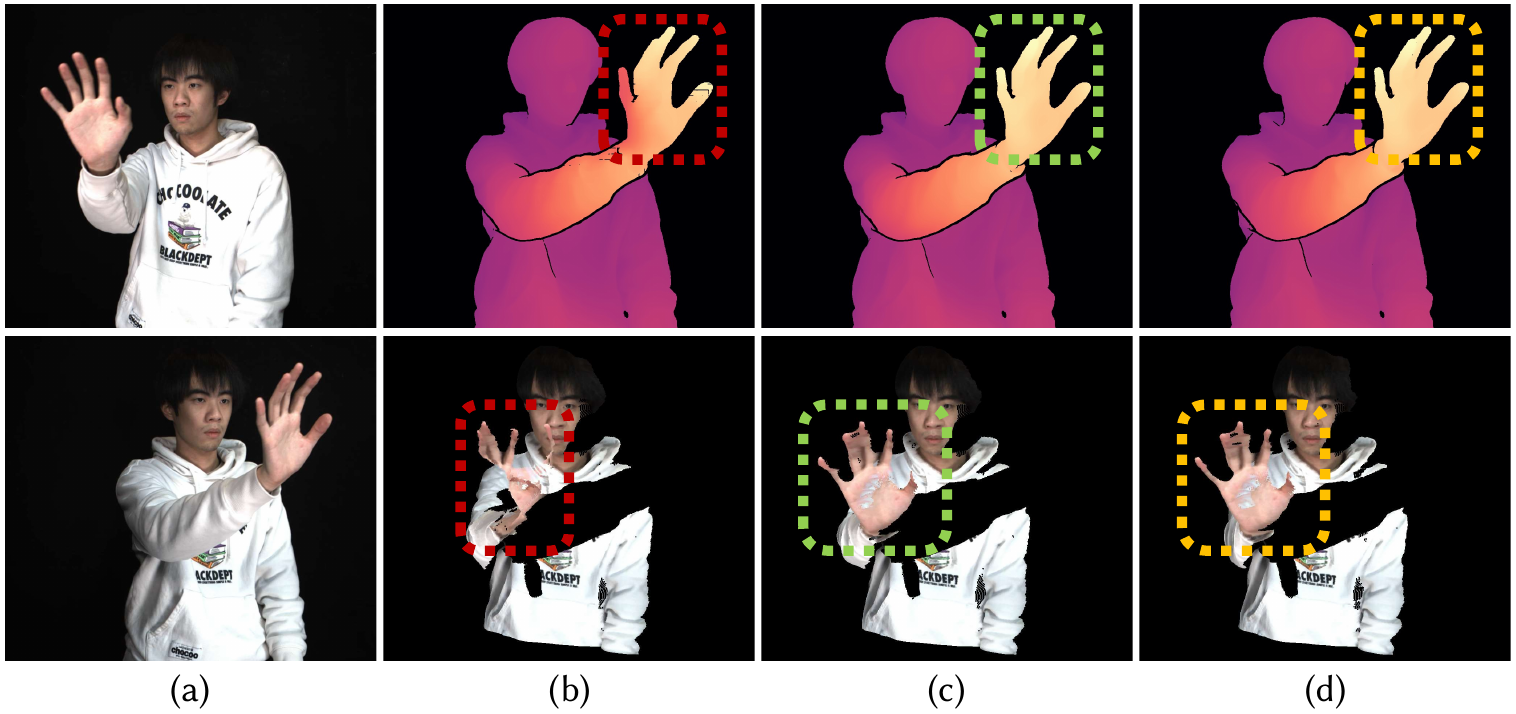}
    \caption{Evaluation of the cascaded disparity estimator. (a) Input views; (b) w/o cascaded initialization, 3 iterative updates; (c) w/ cascaded initialization, 3 iterative updates; (d) w/o cascaded initialization, 16 iterative updates.  }
    \label{fig:casabla}
\end{figure}

For validation of the effectiveness of the cascaded manner, we conduct experiments with synthetic datasets. As shown in Tab. \ref{tab:comp-cascade-disp}, the cascaded initialization improves the accuracy of disparity estimation of large baseline cameras in terms of EPE (end point error) and percentage of small-error pixels. Fig. \ref{fig:casabla} shows that with initialization, 3 iterative flow updates can achieve the same accuracy as 16 iterative updates without initialization, while fewer iterations without initialization cause significant accuracy degeneration.

\begin{figure}
    \centering
    \includegraphics[width=0.95\linewidth]{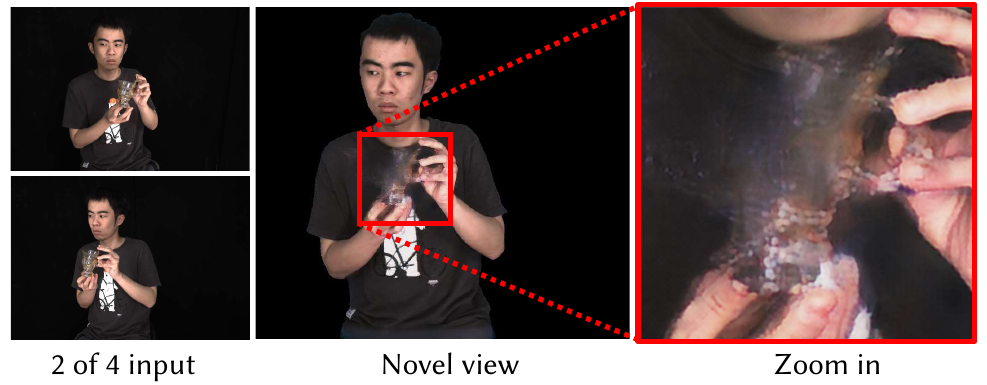}
    \caption{Failure case on non-Lambertian objects.}
    \label{fig:failure_case}
\end{figure}

\subsection{Execution Time and System Latency}

To maximize the performance, we implemented all components carefully with CUDA. Neural networks are optimized using TensorRT with fp16 precision. The overall system is optimized using CUDA Graphs.
The disparity estimation takes 4.7 ms, the encoder takes 6.1 ms, the decoder and the refiner together take 9.3 ms, the input views blending takes 1.4 ms, the 3DGS rasterization takes 1 ms, and other operations take 1 ms, which is in total \textbf{23.5 ms}.

We measured our system’s end-to-end latency in a system containing two terminals in the same LAN.
To determine the latency that users actually sensed, we capture a timer and calculate the time difference between it and the displayed result on the other side. We observe an average latency of around 150 ms, which is acceptable according to \cite{chen2004qos}. We report the detailed latency breakdown of each component of the system in Tab.~\ref{tab:latencyanalysis}.

\begin{table}[htpb!]
    \small
    \caption{Quantitative comparison on synthetic dataset.}
    \label{tab:comp-nvs}
    \begin{tabular}{lccc}
        \toprule
        Method                           & PSNR$\uparrow$  & SSIM$\uparrow$ & LPIPS$\downarrow$ \\
        \midrule
        FloRen~\cite{shao2022floren}     & 19.049          & 0.844          & 0.251             \\
        ENeRF~\cite{lin2022efficient}    & 24.203          & 0.891          & 0.160             \\
        GPS-Gaussian~\cite{zheng2023gps} & 25.204          & 0.909          & 0.112             \\
        Ours w/o refinement              & 25.642          & 0.916          & 0.171             \\
        Ours w/ refinement               & \textbf{26.543} & \textbf{0.928} & \textbf{0.095}    \\
        \bottomrule
    \end{tabular}
\end{table}

\begin{table}[htpb!]
    \small
    \caption{Ablation study on the cascaded disparity module.}
    \label{tab:comp-cascade-disp}
    \begin{tabular}{lcccc}
        \toprule
                                & EPE$\downarrow$ & 1pix$\uparrow$  & 3pix$\uparrow$  & 5pix$\uparrow$  \\
        \midrule
        Small baseline          & 1.141           & 0.8610          & 0.9749          & 0.9861          \\
        Large baseline w/o init & 2.741           & 0.5350          & 0.8302          & 0.9006          \\
        Large baseline w/ init  & \textbf{2.192}  & \textbf{0.7169} & \textbf{0.9051} & \textbf{0.9472} \\
        \bottomrule
    \end{tabular}
\end{table}

\begin{table}[htpb!]
    \small
    \caption{System latency breakdown. }
    \label{tab:latencyanalysis}
    \begin{tabular}{lclc}
        \toprule
        Component       & Latency & Component        & Latency      \\
        \midrule
        Capture         & 56      & RTC transmission & 2            \\
        Upload to vmem  & 11      & Upload to vmem   & 1            \\
        Debayering      & 13      & Decoding         & 5            \\
        Pre-processing  & 5       & View synthesis   & 28           \\
        Encoding        & 5       & Display output   & 27           \\
        Download to mem & 1       & \textbf{Total}   & \textbf{149} \\
        \bottomrule
    \end{tabular}
\end{table}

\section{Conclusion}

We propose a low budget real-time upper-body communication system, Tele-Aloha with only four RGB inputs.
We carefully design a novel view synthesis algorithm for an autostereoscopic displayer, including a cascaded disparity estimation and combination of 3DGS and weighted blending mechanism.
On only one RTX 4090 GPU, we process data capture, stream encoding/decoding, view synthesis and \textit{2K} display with a latency of less than 150 ms.

\paragraph{Limitations} Our system sometimes fails on specular objects, e.g., eyeglasses, due to strong anisotropy which leads to instability of disparity estimation. An example is shown in Fig.~\ref{fig:failure_case}. Also, our method potentially suffers from inaccurate background segmentation, see Fig.~\ref{fig:failure_supp} and for example.

\begin{acks}
    This paper is supported by National Key R\&D Program of China (2022YFF0902200), the NSFC project No.62125107.
\end{acks}

\bibliographystyle{ACM-Reference-Format}
\bibliography{bibliography}

\appendix

\begin{figure*}
    \centering
    \includegraphics[width=.85\linewidth]{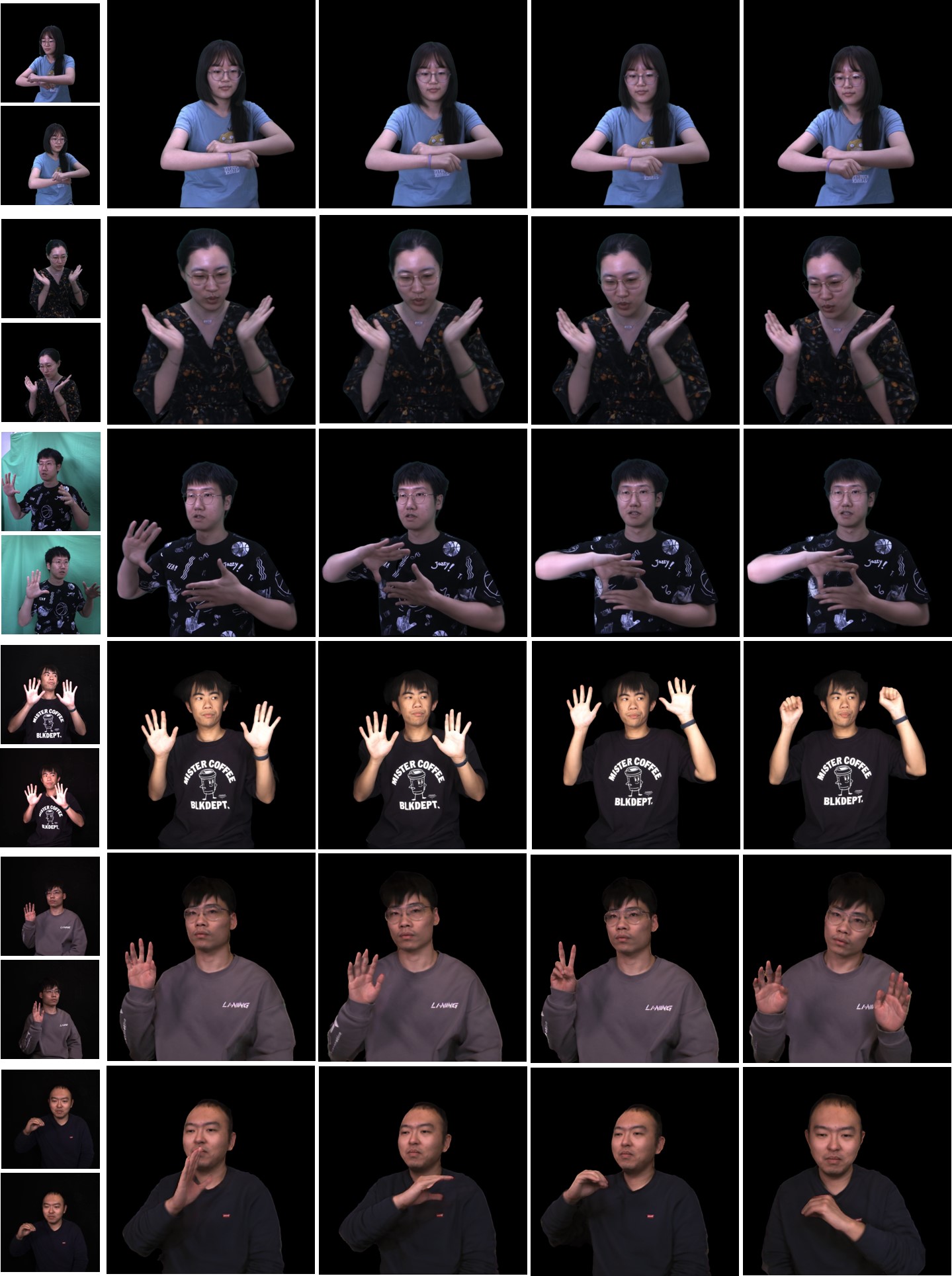}
    \caption{Main results of Tele-Aloha system. The first column shows 2 of 4 input views. The first two rows show the results of static scenes. The other rows show the results of dynamic scenes. }
    \label{fig:mainresults}
\end{figure*}

\begin{figure*}
    \centering
    \includegraphics[width=0.99\linewidth]{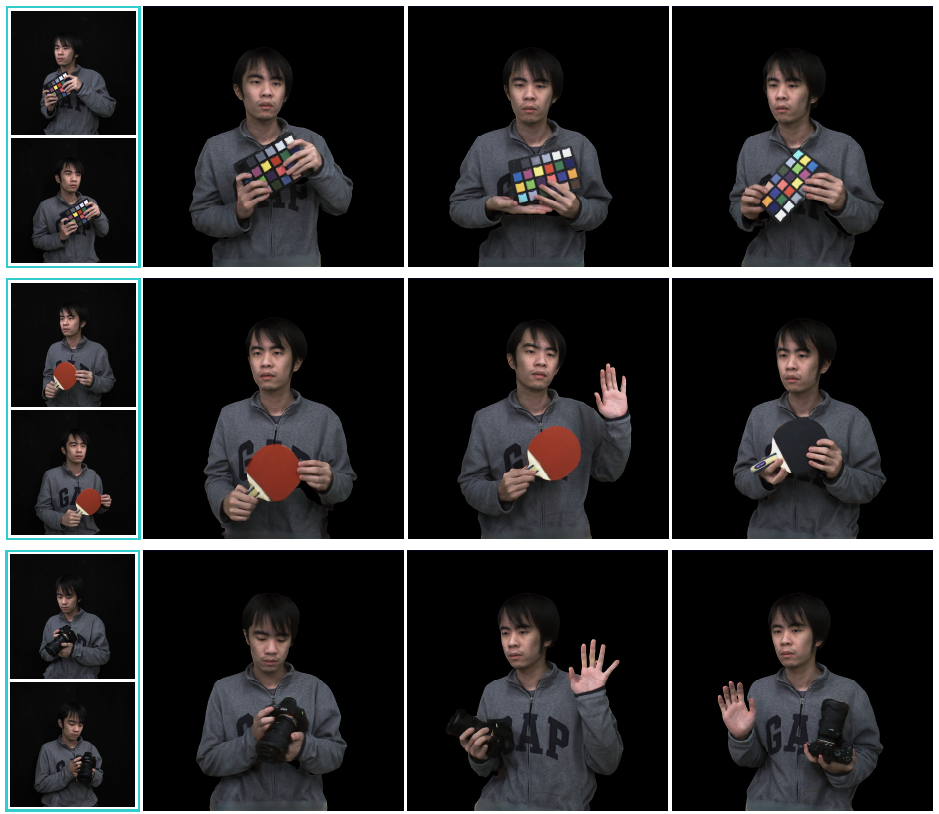}
    \caption{Results of Tele-Aloha system on human-object interaction scenes. Our method can synthesis most of solid objects.}
    \label{fig:objects}
\end{figure*}

\begin{figure*}
    \centering
    \includegraphics[width=0.99\linewidth]{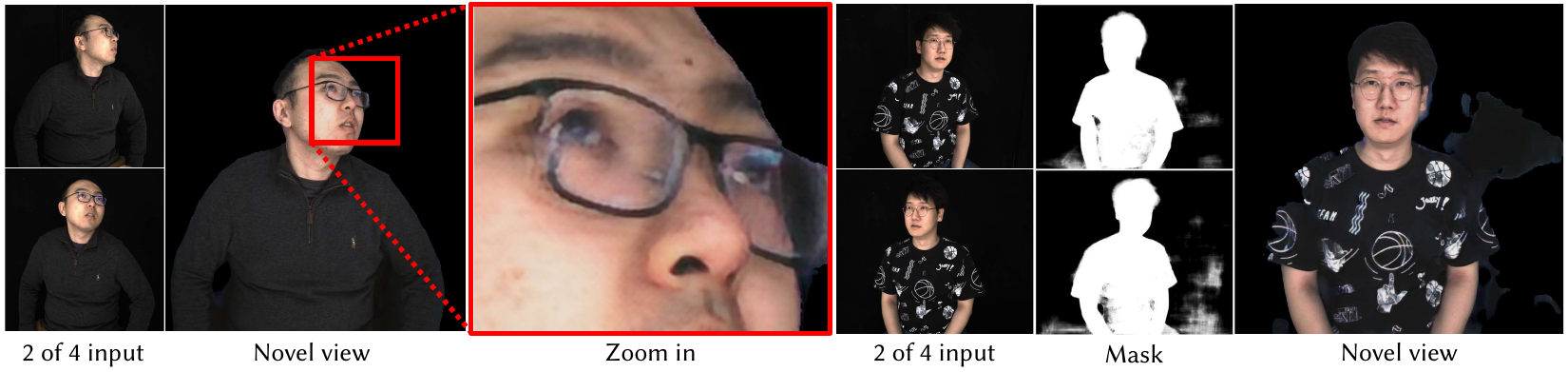}
    \caption{Failure cases. Left one shows blurry eyeglasses. Right
        one shows that inaccuracy of background matting causes artifacts in novel views.}
    \label{fig:failure_supp}
\end{figure*}

\end{document}